\documentclass{INTERSPEECH2023}

\interspeechcameraready 

\usepackage{cite}
\usepackage{tikz}
\usepackage{qtree}
\usepackage{verbatimbox}
\usepackage{amsmath,amssymb,amsfonts}
\usepackage{algorithmic}
\usepackage{graphicx}
\usepackage{textcomp}
\usepackage{xcolor}
\usepackage{tikz}
\usetikzlibrary{trees,positioning,fit}
\usepackage[utf8]{inputenc}
\usepackage[T1]{fontenc}
\usepackage{lmodern}
\usepackage{multirow}
\usepackage{booktabs}
\usepackage{color, colortbl}
\usepackage{tabularx, float, makecell}

\newcommand{\keypoint}[1]{{\newline}{\noindent}{\textbf{#1}}}

\newcommand{\model}{BeAts }
\usepackage{hyperref}
\hypersetup{
  colorlinks=true,
  linkcolor=black,
  urlcolor=black,
  hidelinks=true
}

\renewcommand{\arraystretch}{1.5}
\newcolumntype{s}{>{\columncolor[HTML]{E6EFFC}} p{2.5cm}}
\arrayrulecolor[HTML]{97B3D9}
\def\BibTeX{{\rm B\kern-.05em{\sc i\kern-.025em b}\kern-.08em
    T\kern-.1667em\lower.7ex\hbox{E}\kern-.125emX}}
    
\title{\textsc{BeAts}: Bengali Speech Acts Recognition using Multimodal Attention Fusion}

\name{Ahana Deb$^{\ast 1}$, Sayan Nag$^{\ast 2}$, Ayan Mahapatra$^{\ast 1}$, Soumitri Chattopadhyay$^{\ast 1}$, Aritra Marik$^{\ast 1}$, Pijush Kanti Gayen$^1$, Shankha Sanyal$^1$, Archi Banerjee$^3$, Samir Karmakar$^1$}
\address{
  $^1$Jadavpur University, India \qquad $^2$University of Toronto, Canada \qquad $^3$IIT Kharagpur, India}
\email{ahanadeb01@gmail.com, nagsayan112358@gmail.com, ssanyal.ling@jadavpuruniversity.in}

\begin{document}

\maketitle
 
\begin{abstract}
   Spoken languages often utilise intonation, rhythm, intensity, and structure, to communicate intention, which can be interpreted differently depending on the rhythm of speech of their utterance. These speech acts provide the foundation of communication and are unique in expression to the language. Recent advancements in attention-based models, demonstrating their ability to learn powerful representations from multilingual datasets, have performed well in speech tasks and are ideal to model specific tasks in low resource languages. Here, we develop a novel multimodal approach combining two models, wav2vec2.0 for audio and MarianMT for text translation, by using multimodal attention fusion to predict speech acts in our prepared Bengali speech corpus. We also show that our model \model(\underline{\textbf{Be}}ngali speech acts recognition using Multimodal \underline{\textbf{At}}tention Fu\underline{\textbf{s}}ion) significantly outperforms both the unimodal baseline using only speech data and a simpler bimodal fusion using both speech and text data.
   Project page: \href{https://soumitri2001.github.io/BeAts}{\color{blue}{https://soumitri2001.github.io/BeAts}}
   \keypoint{Index Terms:} speech act, multimodal fusion, transformer, low-resource language
\end{abstract}
\noindent

\def\thefootnote{$\ast$}\footnotetext{Equal contribution}\def\thefootnote{\arabic{footnote}}

\section{Introduction}

According to the Speech Act theory\cite{Speech-Act-Theory}, issuance and utterance of words, which happens during the articulation of speech, provide the foundation of communication between the speaker and the listener. In any communication through spoken language, the listener is dependent on their own ability to decode the explicit or implicit intention encoded in the speaker's delivery for the communication to be successful. Every spoken language has its unique set of morphemes, intonations, and sentence structures, that adds crucial meaning to the message being conveyed. Unique arrangements of these building blocks of speech can completely change the proper meaning intended by the speaker. Thus, the same utterance of a sentence can be interpreted differently, depending completely on the rhythm of speech of utterance. 

\begin{figure}[!ht]
    \centering
    \includegraphics[width=0.45\textwidth]{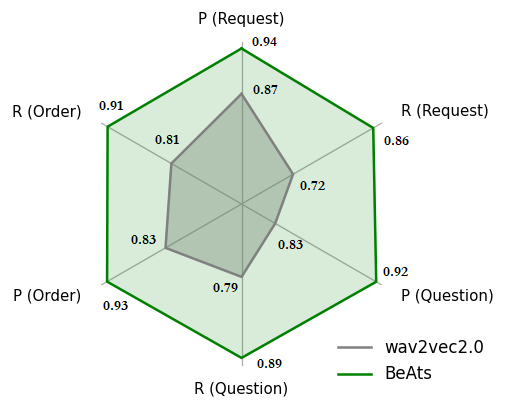}
    \caption{\textnormal{\textbf{BeAts}} \textbf{performs better than unimodal wav2vec2.0} across all 3 speech acts classes in both Precision (P) and Recall (R) scores, by learning rich features utilizing multimodal attention fusion module.}
    \label{fig:abstractfig}
\end{figure}

Prior studies in speech act recognition relied on both linguistics and non-linguistic aspects, like sequential context, physical context, adjacency pair, etc  \cite{Levinson, Gisladottir}. However one of the critical aspects of speech recognition in multilayered speech act condition is prosody and intonation, and the modulation of parameters of this prosody changes the meaning of the intended communication. For instance, a simple statement like "The sun rises in the east", can be interpreted as a statement, but also with a slight change by raising the tone in the last phrase uttered, can imply that the speaker intends it as a question. Therefore, with the same morpho-syntactic strings, a difference in intonation allows the same sentence to be interpreted as a question or an assertion. In this study we make an attempt to explore this domain for Bengali (Bangla), a low-resource Indo-Aryan language.

Our contributions are: (i) preparing a corpus of speech acts in Bengali \footnote{Dataset can be found in \href{https://soumitri2001.github.io/BeAts}{\color{blue}{Project Page}}}, (ii) proposing a novel multi-modal architecture (baseline) and investigating its performance in classifying these speech acts in a low-resource setting.

\section{Related works}

Initial works in processing and classification of speech data rely mostly on classifying low level acoustic speech parameters and spectral features using SVMs, and shallow NNs \cite{Scherer,Scherer2, Bach, Tao, Verv, Lech}. Further, CNNs were used for emotion recognition task in speech \cite{Mao, Fayek}. For speech act classification task, Kipp, 1998 \cite{Kipp} trained an Elman/Jordan network on German speech. Furthermore, another study \cite{Marineau} used neural network, parsing model, and a linguistic rule-based classifier to classify speech acts of Assertion, WH-Question, Directive, and Yes/No Questions. 

To utilise the various aspects of human communication to decode the underlying meaning, multimodal structures were introduced. Earlier works include \cite{Chuang} which proposed the use of SVMs for classifying the acoustic features forming a feature subspace, and manually defining emotional keywords for text analysis. Subsequent works on multimodal analysis followed partial reliance on manufactured features, using ConvNet structures to process both textual and visual features \cite{Poria}. Yue Gu et. al, 2017\cite{Yue} presented a multimodal structure consisting of two independent CNNs, processing speech and text data. 

Recent success of transformers in sequence modeling tasks has seen a wide application in automatic speech recognition, which completely eliminates the need to create pre-defined features. In this regard, BERT-like \cite{BERT} models achieved a significant improvement on previous architectures. Following BERT, subsequent architectures emerged for audio data modeling, such as the wav2vec2.0 model\cite{wav2vec2}. During pre-training the model learns general representations of speech phonemes and complex features, providing a good starting point to be fine-tuned on other low resource languages, such as Bengali, with labeled data, for speech acts classification task. 

In our approach, we use pretrained wav2vec2.0, and fine-tune the model to classify speech acts in Bengali speech audio. We further fine-tune Marian-NMT\cite{mariannmt}, a neural machine translation framework pre-trained on multilingual datasets\cite{opus-mt}, on Bengali text to English text transcription of the speech data. We use a multimodal-modal attention fusion using two separate schemes, firstly with optimal transport kernels\cite{otk} and secondly with multimodality transformer. This facilitates a greater level of interconnectivity and fusion between the two networks processing different data from different modalities. Finally, we feed the output to a series of fully connected layers and a decision softmax layer for prediction.

In contrast to several recent works on multimodal speech and text processing, we use self-attention heads to learn both speech and text latent representations, and feed them to a separate downstream model. We introduce a novel architecture BeAts, and extend the work of classifying perceptually understood speech acts into our prepared Bengali speech corpus, and we compare our architectures with the individual unimodal and bimodal models evaluated on the same dataset.

\section{Experimental Setup}
\begin{figure}[!h]
    \centering
    \includegraphics[width=0.45\textwidth, height=0.25\textwidth]{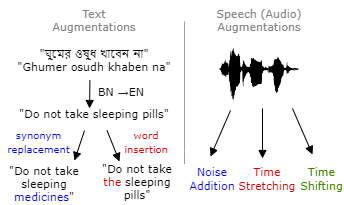}
    \caption{Augmentations used in this study.}
    \label{fig:augmentation}
\end{figure}

We prepared a dataset of 85 utterances, consisting of request, question, and order, having 25,35, and 25 utterances in the class groups respectively. The duration of the chosen utterances were approx 1300 ms and with 5 to 7 words each recorded in 44.1 kHz sampling rate in a soundproof recording room. For our experiment 2 male and 2 female (average age = 23 years, SD= 2,3 years) L1 Bengali native speakers were chosen. Prior to the experiment, the participants were instructed to carefully read each of the sentence and comprehend their meanings. The sentences for each individual category were recorded one after another without providing any prior context.
A standard amplitude normalization protocol was used after the recording.
Since the number of samples in the dataset is small we resorted to using augmentations (to increase effective sample size) as shown in Fig \ref{fig:augmentation}. Here we refrained from doing any spectral augmentations in the audio dataset because that might have an adverse effect on the data. Further, we did not augment Bengali text because the parity between the audio and the text will be lost. However, we carefully augmented the English samples so that it prevents the semantic meaning upon translation.





\section{Methodology}

\begin{figure*}[t]
    \centering
    \includegraphics[width=\textwidth,height=0.5\textwidth]{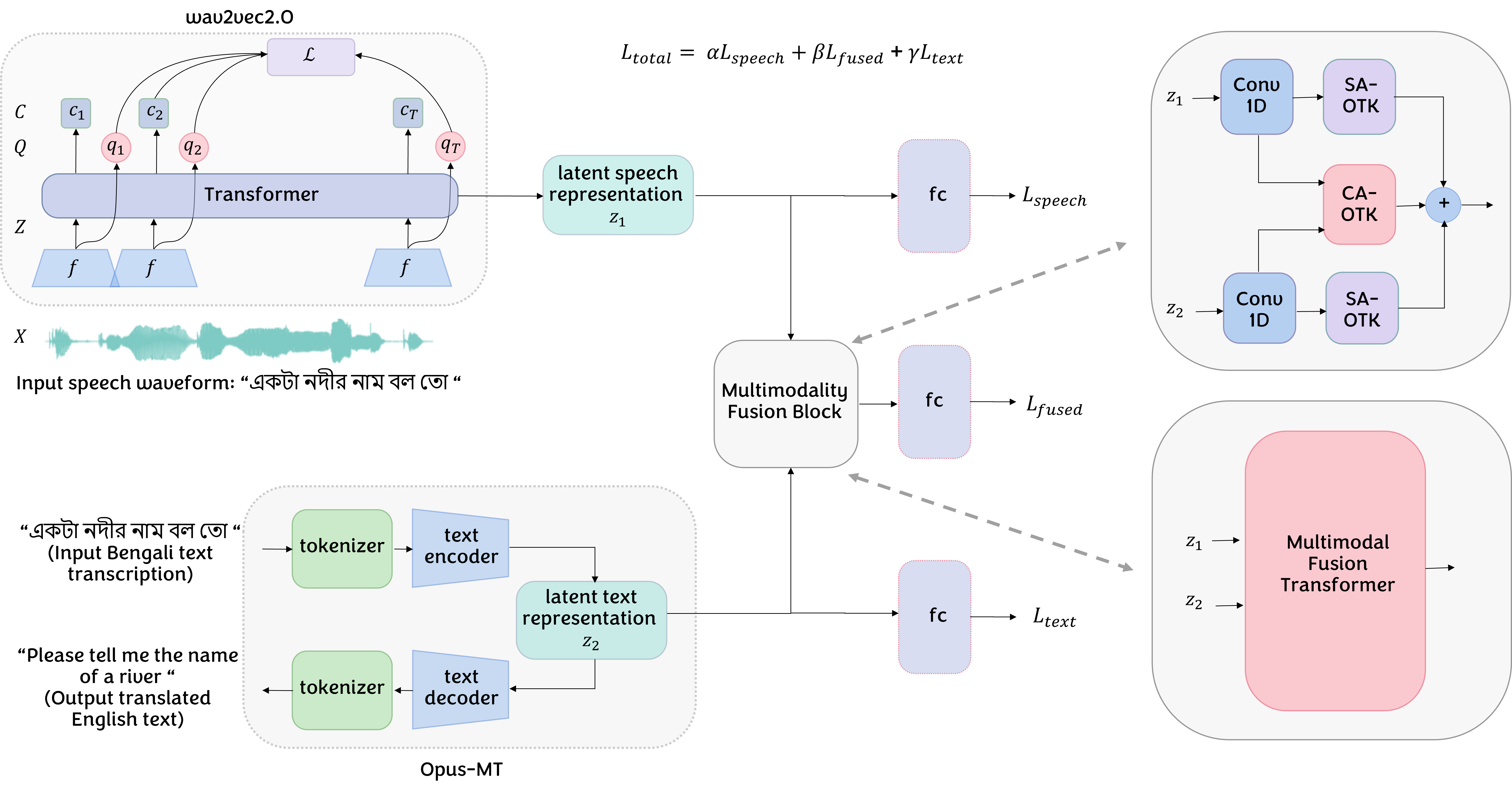}
    \caption{\model takes in as input raw waveform, annotated with the speech act class labels, and transcribed Bengali to English translation text input. The data undergoes positional embedding and is fed as input to the transformer architectures for sequence modeling task. The respective outputs are fed to a multimodality fusion block comprising of two separate schemes (i) an optimal transport kernel (OTK) based attention, and (ii) a multimodal fusion transformer. The output of this fusion block is passed through fully connected layers for classification task.}
    \label{fig:architecture}
\end{figure*}

\subsection{wav2vec2.0}

Raw wave input, which is positionally encoded by a convolutional neural network (CNN), GELU activation function\cite{gelu} and layer normalised, is provided as an input to the wav2vec2.0\cite{wav2vec2} transformer architecture as shown in Fig \ref{fig:architecture}.
We use wav2vec2.0 which has been pre-trained on Librispeech and LibriVox dataset using contrastive loss (CL). CL is obtained by computing the cosine similarity between the context representation (obtained at the transformer output) and the quantized representation (produced through choosing discrete entries from codebooks). 
During finetuning this model on our audio dataset, the final layer is modified to output probability scores of the classes, and a binary cross-entropy loss is calculated for upgrading gradients.


    
    
    
    

    
    
    
    

\subsection{Opus-MT}

Here, we used contextual text representation obtained in Bengali to English translation, as an additional input for the classification task.  This is motivated by the clear structural distinction in how requests are expressed from orders in English, which is not observed in Bengali. For example, the sentence “Ekta nodir naam bolo toh” has the same utterance for both question and request in Bengali, however in English, it is expressed as “Can you tell me the name of a river” when it is intended as a question, and as “Please tell me the name of a river” when intended as a request. Our goal is to utilise this difference in interjections in these two speech act classes in English, to improve our performance on classifying speech acts in Bengali, by learning a latent representation of translation from Bengali to English of this transcribed text.



\begin{figure}[!h]
    \centering
    \includegraphics[width=0.42\textwidth]{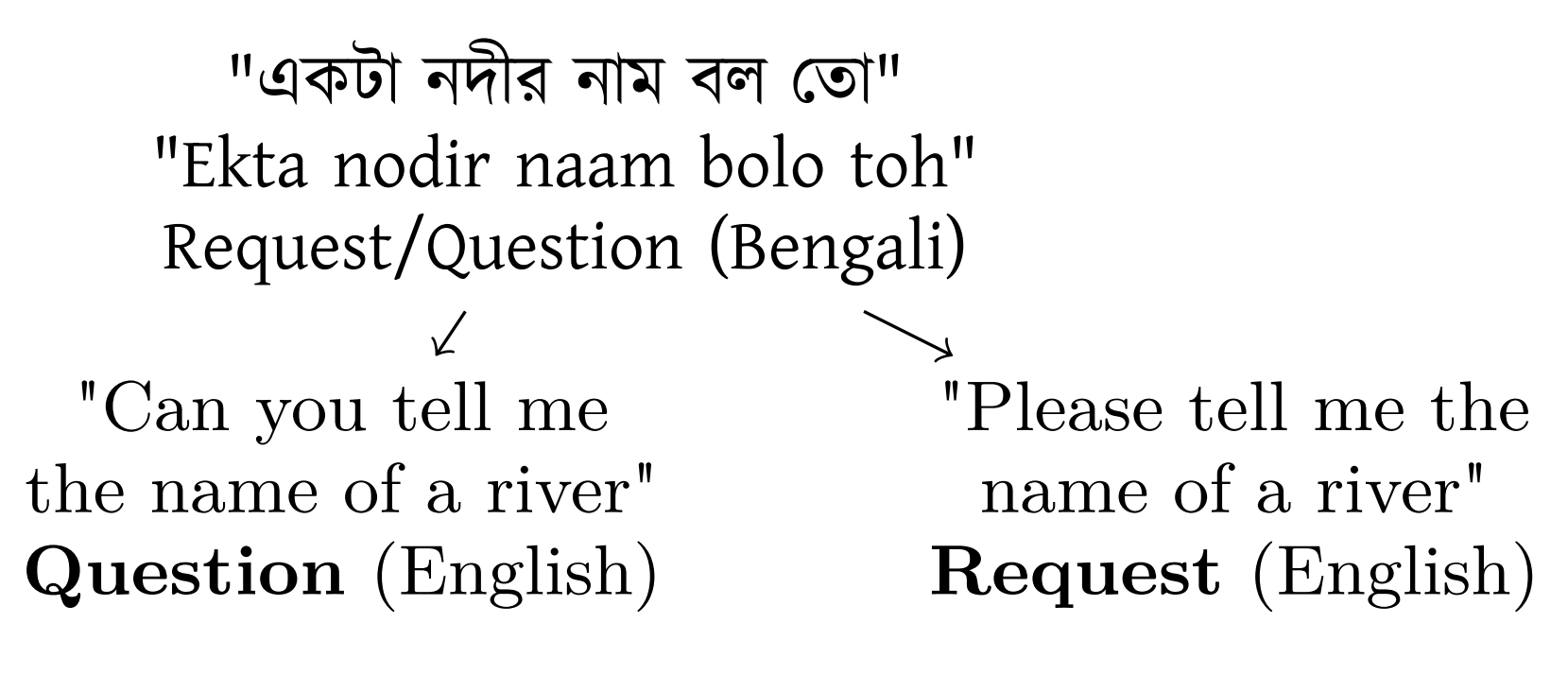}
    \caption{The same utterance can be interpreted as Request or Question in Bengali, whereas the expressions are structurally different in English.}
    \label{fig:my_label}
\end{figure}

The Opus-MT model\cite{opus-mt} utilises Marian-NMT(Neural Machine Translation)\cite{mariannmt} as its framework architecture, based on the standard transformer architecture.
The model is pre-trained on large bitext repository OPUS\cite{tiedemann-2012-parallel}, and fine-tuned on the contextually transcribed Bengali to English text of our Bengali speech data set for the speech act classification task. The architecture follows the base model described in \cite{vaswani}. Here, the latent representation is concatenated with the latent speech representation obtained from the wav2vec2.0 model, and fed to fully connected layers to output a softmax prediction over the individual classes.

\begin{figure*}[!h]
    \centering
    \includegraphics[width=\textwidth]{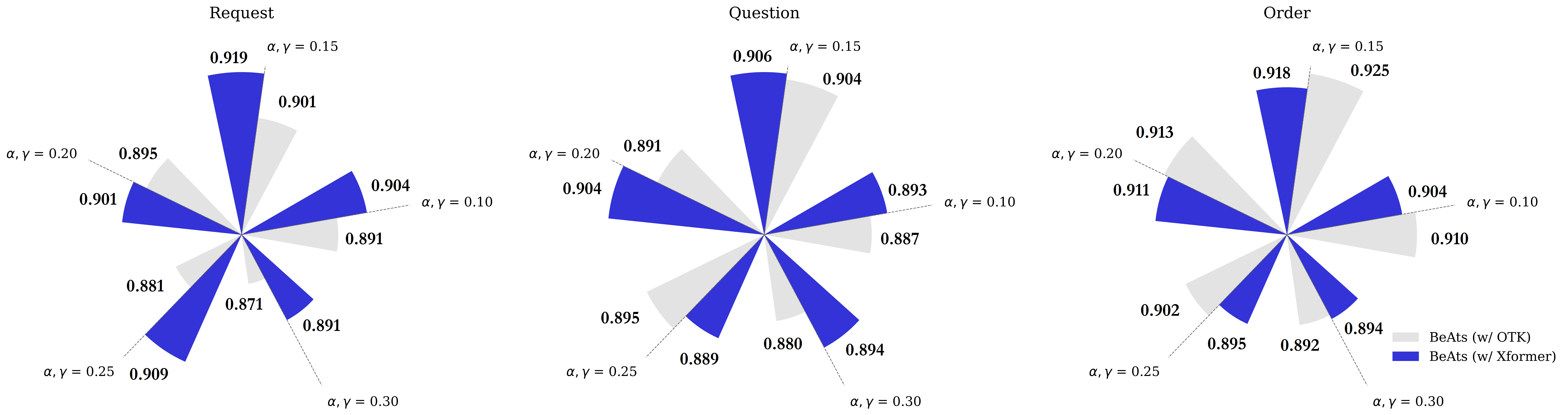}
    \caption{Ablation with F1 scores.}
    \label{fig:ablation}
\end{figure*}

\subsection{Multimodal Attention Fusion}

To further better our model’s performance, instead of directly concatenating the latent speech representation and text translation representation from the two models, we feed them separately into a multimodality fusion block consisting of two schemes, namely, an Optimal Transport Kernel (OTK) scheme, and a multimodal fusion transformer scheme. In both these schemes, we use the output of the multimodality fusion block and pass it through fully connected layer. Furthermore, for training we introduce a joint loss combining three weighted loss terms (Fig \ref{fig:architecture}):
\begin{equation}
    L_{total} = \alpha L_{speech} + \beta L_{fused} + \gamma L_{text}
\end{equation}
Ablation studies on these weights are discussed in the following subsection.

\textbf{Multimodal Fusion Transformer:} Here, we propose a MultiModal Fusion Transformer where we adapt transformers for fusion among the speech and text modalities. We concatenate the features from respective modalities together along with a special [CLS] token as the first feature vector of the agregated sequence to be used as input to the multimodal transformer.

\textbf{OTK:} The intuition of choosing OTKs is their robustness which have been clearly demonstrated over usual aggregation methods (mean/max pooling or attention) in recent studies, in long sequences of varying sizes, with long range dependencies. Furthermore a single layer of OTKs have also outperformed multi-layer feed forward NNs and in some cases multi-layer CNNs too\cite{otk}. Learning representations from a large set of feature vectors with long range interactions actively benefits from pooling to reduce the number of feature vectors, and here the pooling rule is similar to that of self-attention as it follows inductive bias, aggregating features based on similarity scores. Given a set and a learned reference, alignment of these elements are done using optimal transport protocol, and then using this similarity they are weighted accordingly to be pooled, which produces the output embedding.
Cross attention is an intuitive multimodal fusion method in attentive modules where attention masks from one modality (audio representations) are used to highlight the extracted features in other modalities (text representations). This is different from self-attention where attention masks from one modality (text) are used to highlight it’s own features. We use a combination of both cross attention (multimodal attentive fusion) and self-attention to facilitate a greater level of interconnectivity and fusion between the two networks processing different data from different modalities, while also preserving the individual representations of the models via the respective self-attention layers.

\begin{table}[!h]
    \centering
    \resizebox{1\columnwidth}{!}{\begin{tabular}{l ccc ccc ccc}

    \toprule
    
    \textbf{Model} & \multicolumn{2}{c}{\textbf{Request}} & \multicolumn{2}{c}{\textbf{Question}} & \multicolumn{2}{c}{\textbf{Order}}\\[0.5ex]
    \cmidrule(lr){2-3}
    \cmidrule(lr){4-5}
    \cmidrule(lr){6-7}
    
     & \textbf{P} & \textbf{R} & \textbf{P} & \textbf{R} & \textbf{P} & \textbf{R} \\[0.5ex]
    \midrule
    
    
    wav2vec2.0  & 0.87 & 0.72 & 0.83 & 0.79 & 0.83 & 0.81 \\[0.5ex]

    
    \multirow{2}{*}{wav2vec2.0}\\
     +Opus-MT & 0.91 & 0.73 & 0.86 & 0.80 & 0.88 & 0.82\\[0.5ex]

    \rowcolor[HTML]{E6EFFC}
     \model(Xformer) & \textbf{0.94} & \textbf{0.89} & 0.91 & \textbf{0.90} & \textbf{0.93} & 0.90\\[0.5ex]

    \rowcolor[HTML]{E6EFFC}
    \model(OTK) & \textbf{0.94} & 0.86 & \textbf{0.92} & 0.89 & \textbf{0.93} & \textbf{0.91}\\[0.5ex]
    
    \bottomrule
    \end{tabular}}\\[0.5ex]
    \caption{Precision (P) and Recall (R) scores of the respective speech act classes.}
    \label{tab:1}
\end{table}

\vspace{-2.4em}

\section{Results}
As a baseline, we initially performed unimodal classification with wav2vec2.0. The Precision (P) and Recall (R) scores are reported in Table \ref{tab:1}. Based on the results, it is quite evident that the speech modality alone is not enough to capture all the information. Therefore, given the assumption that Question-Request speech act expression is structurally distinct in English and Bengali, we further fine-tune Opus-MT transformer on our transcribed dataset, and concatenate the two latent representations obtained from the models, and use fully-connected layers to produce a softmax score. This bimodal approach increased the overall performance as compared to the unimodal approach, as can be ascertained from Table \ref{tab:1}. 

The experimental results for \model are listed in Fig \ref{tab:1}. We extend our bimodal approach by using multimodal attention fusion via OTKs and Multimodal Transformers respectively. \model achieved a significant boost in performance across all classes for both the schemes \ref{tab:1} indicating the impact of multimodal fusion in the Bengali speech acts classification task.

Furthermore, we did ablation studies on the joint loss function (Eq. 1) used for training the \model model. For simplicity, we have considered $\alpha = \gamma$ and therefore, $\beta = 1 - 2\alpha = 1 - 2\gamma$. We have considered $\alpha = \gamma = [0.1, 0.15, 0.20, 0.25, 0.30]$ and reported the F1 scores for both OTK and Multimodal Transformer (Xformer) schemes for all the 3 classes (Fig \ref{fig:ablation}). From our experiments, we have observed that the best performance is obtained for a value of 0.15.

\setlength{\arrayrulewidth}{.1mm}
\setlength{\tabcolsep}{5pt}
\renewcommand{\arraystretch}{1.7}
\newcolumntype{s}{>{\columncolor[HTML]{E6EFFC}} p{2.5cm}}
\arrayrulecolor[HTML]{97B3D9}










\section{Conclusion}

In this work we presented a novel multimodal approach for speech act classification in Bengali utterances. We combine evidence from the intonations of the utterances as well as the structural difference in speech act classes in English, to obtain significantly better performance on our Bengal speech corpus, compared to the individual unimodal and bimodal approaches. Here a wav2vec2.0 transformer models multilingual speech data and a Marian-NMT transformer models neural machine translation, and the combination of these two significantly increases the accuracy of speech act classification from just using a single wav2vec2.0 model using only audio data to classify. Furthermore, the performance of multimodal attention fusion is proved to be much better than solely using a fully-connected layer in combining latent space representations. Both wav2vec2.0 based audio transformers and Marian-NMT like text transformers are demonstrated as multilingual models, which after being pre-trained in other widely used (or with availability of rich datasets) languages, can be applied to more diverse and data/resource constrained languages directly, without the loss of nuance.

Our novel method of using both learning from audio data and text data to classify speech acts on our prepared Bengali speech corpus demonstrates the usefulness and robustness of human-like multimodal approach to learning and navigating specifically prosodic nuances and broadly any other language specific spoken communication. Our future works will include further improvements on this study by updating the dataset and applying novel architectures. We are excited about the future of multimodal language processing, especially in a low-resource setting, where a limited amount of labeled data can improve the performance of a classification task, used to produce more accurate translation, eliminate specific inherent biases taken in from data.



\bibliographystyle{IEEEtran}

\bibliography{mybib}

\end{document}